\pdfoutput=1
\documentclass[a4paper]{article}

\usepackage{INTERSPEECH2022}

\title{Enhancing Speech Recognition Decoding via Layer Aggregation}
\name{Tomer Wullach, Shlomo E. Chazan}
\address{OriginAI, Israel}
\email{tomerw@originai.co, shlomi@originai.co}
\begin{document}

\maketitle
\begin{abstract}
  Recently proposed speech recognition systems are designed to predict using representations generated by their top layers, employing greedy decoding which isolates each timestep from the rest of the sequence. Aiming for improved performance, a beam search algorithm is frequently utilized and a language model is incorporated to assist with ranking the top candidates.
  In this work, we experiment with several speech recognition models and find that logits predicted using the top layers may hamper beam search from achieving optimal results. Specifically, we show that fined-tuned Wav2Vec 2.0 and HuBERT yield highly confident predictions, and hypothesize that the predictions are based on local information and may not take full advantage of the information encoded in intermediate layers. To this end, we perform a layer analysis to reveal and visualize how predictions evolve throughout the inference flow. We then propose a prediction method that aggregates the top $M$ layers, potentially leveraging useful information encoded in intermediate layers and relaxing model confidence. We showcase the effectiveness of our approach via beam search decoding, conducting our experiments on Librispeech test and dev sets and achieving WER, and CER reduction of up to 10\% and 22\%, respectively. 
\end{abstract}
\noindent\textbf{Index Terms}: Self-supervised learning, speech recognition, representation analysis

\section{Introduction}

Self-Supervised Learning (SSL) has been widely adopted in a variety of tasks~\cite{devlin2018bert, liu2019roberta, liu2020multilingual, baevski2020wav2vec, liu2020mockingjay} and shown to be capable of producing representation which can be effectively utilized for downstream tasks. Models trained via SSL rely on information learned during what is referred to as pre-training step, leveraging massive amount of unlabeled data, and further tuned for specific tasks using significantly smaller amount of labeled data. In recent years, SSL has been employed for speech processing tasks such as Automatic Speech Recognition (ASR)~\cite{baevski2020wav2vec, hsu2021hubert}, Phoneme Classification (PC)~\cite{liu2020mockingjay, baevski2020wav2vec, van2018representation}, Speaker Recognition (SR)~\cite{liu2020mockingjay} and others, utilizing transformer encoder layers~\cite{vaswani2017attention} yielding context aware representations. 
 
Recently proposed ASR models use CTC~\cite{graves2006connectionist} variants for both training and decoding, achieving state-of-the-art results~\cite{baevski2020wav2vec, hsu2021hubert}. A traditionally used decoding strategy would be selecting the token with the highest predicted probability at each timestep. However, this greedy decoding approach is sub-optimal as greedy methods might disregard sequences of higher quality. Seeking to break away from the greedy approach which isolates the prediction at each timestep, there has been several attempts to incorporate a transformer decoder layers in speech processing models~\cite{deng2021improving, morais2021end, kanda2021large} showing noticeable performance improvement. Yet, employing an autoregressive decoder compromise the model's ability to parallelize processing procedures. To this end, many works maintain a search-based decoding scheme such as beam search~\cite{baevski2020wav2vec, chen2021wavlm, hsu2021hubert}. Additional gains are likely to be obtained when an external language model (LM) is integrated within the decoding process~\cite{hsu2021hubert, baevski2020wav2vec, chen2021wavlm, gulati2020conformer}.

As opposed to greedy search algorithms, beam search generate multiple transcription candidates for an input sequence by considering several alternatives at each timestep. At each timestep, beam search rank all possible classes using their predicted probabilities and keep the $B$ token with highest probability, where $B$ is the beam width. Although enabling easy, straightforward decoding, beam search might be prone to biases when a model's predictions are highly confident. In that case, the predicted probability distribution at each timestep is massed to a single token, reducing the relevancy of the beam width as very few candidate sequences dominate all others. As a consequence, this case of highly confident predictions prevent the algorithm from truly considering multiple alternatives. For this purpose, we monitor and perform a layer analysis of the confidence of several high performing ASR models. 

Our findings regarding the behaviour of the transformer layers during inference reveal high levels of confidence in the top layers. To tackle this issue we propose aggregating the logits of the top $M$ transformer layers, and use the aggregated logits as the beam search input. From another perspective, previous works has shown that different layers encode different aspects of the data~\cite{pasad2021layer} such as linguistic information and acoustic features. Thus, predictions based on representations produced by the final layer may neglect informative features or give excessive weight to others. Recent studies has shown the merits of aggregating deep transformer layers~\cite{yang2020novel}, demonstrating that aggregating representations encoding different features creates an enriched representation that boost performance~\cite{dou2020exploiting, wang2020multi}.

In this work, we suggest aggregating the logits predicted by the last $M<N$ transformer layers, creating modified relaxed logits that are: (1) less confident and more diverse, providing beam search a larger pool of alternatives. (2) include linguistic and acoustic information which is possibly lacking from representations created solely using the top layer. To the best of our knowledge, this is the first attempt to employ layer aggregation of pre-trained SSL-based models for speech recognition. Our contributions can be summarized as follows:
\begin{enumerate}
    \item We conduct an empirical study revealing the layer-wise confidence levels of SSL-based speech recognition systems.
    \item We showcase the findings from 1. and visualize the impact on the predicted transcriptions when decoding using a beam search.
    \item We propose a layer aggregation method that harness linguistic and acoustic features encoded in intermediate layers and relaxes confidence of predicted logits.
\end{enumerate}

Extensive experimental studies show that the proposed method, although not including additional training, is beneficial and reduces the WER and CER with up to $10\%$ and $22\%$, respectively.   

\begin{figure*}[th]
  \includegraphics[trim=0 90 0 20, clip,width=\textwidth]{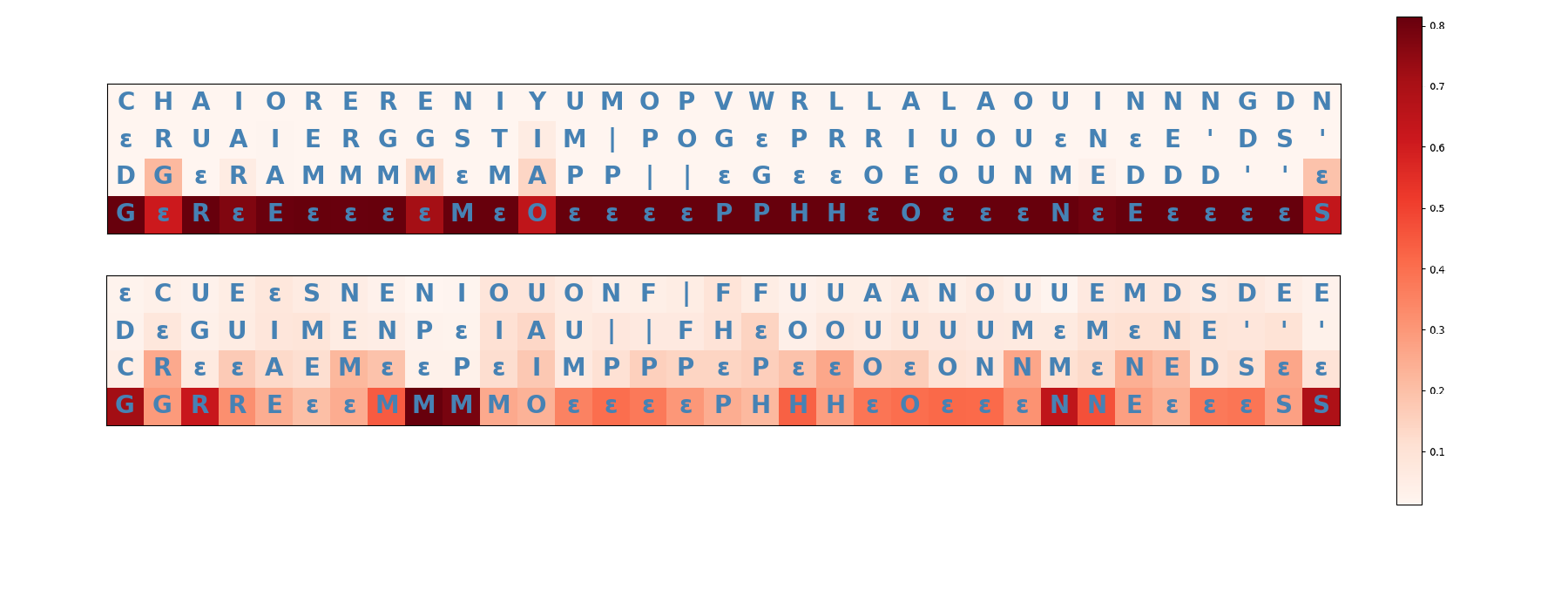}
  \begin{center}
    \caption{Example from Librispeech test-clean: \emph{"...people are not so different from \textbf{gramophones}..."}.Top 4 probabilities for each timestep, predicted using wav2vec 2.0 Large fine-tuned using 960 hours of Librispeech data. The model tends to produce exceptionally confident predictions (upper figure). The proposed layer aggregation approach (lower figure) was able to correctly predict \emph{"gramophones"} using beam search by relaxing the model's confidence, and diversifying the considered candidates during decoding.} 
    \label{fig:logits}
  \end{center}
\end{figure*}

\section{Prediction Flow}
\subsection{Speech Recognition Inference}
Let $M$ be a trained speech recognition model, which was pre-trained using SSL and fine-tuned for speech recognition on dataset $D$. Composing $M$ are a feature extraction module, and a contextual module with $N$ transformer encoder layers~\cite{baevski2020wav2vec, hsu2021hubert}. We denote $H^{t}_n$ as the representation produced by transformer layer $H_n$ at timestep $t$. Let $lm\_head(H_N)$ be a linear projection layer, projecting emissions from the top transformer layer $H_N$ into $C$ classes, where $C$ is the number of possible tokens (i.e., vocabulary size).

In this work, we investigate SSL pre-trained models fine-tuned using the objective of minimizing a CTC loss~\cite{graves2006connectionist}. Thus, for a raw input audio sequence $X$, the feature extraction module extracts latent representations, resulting with a sequence of speech representations $Z$ of length $T$. $Z$ is then fed to the contextual module which predicts a logit matrix of shape $\mathbb{R}^{T \times C}$. The goal is to estimate the sequence $Y=\left[ y_1,\ldots,y_{T}\right]$, given input sequence $X$. Hence, the acoustic model output is defined with, $P_{\text{AM}}(Y|X) \in \mathbb{R}^{T \times C}$.

\subsection{Greedy and Beam Search Decoding}

At each timestep, the token selection heuristic aim to maximize the probability  $P(Y|X)$, as described in Equation~\eqref{eq:argmax}.

\begin{equation}
  P_g(Y|X) = \underset{Y}{\arg\max} \prod_{t = 1}^{T} {p_{t}(y_t | X)}
  \label{eq:argmax}
\end{equation}

However, generating predictions using this kind of greedy decoding might neglect candidates with greater likelihood which consist of token that does not necessarily attain the maximum probability at each timestep.
For this purpose, a beam search is frequently employed, combining both logits and LM scores as described in Equation~\eqref{eq:lm}:

\begin{equation}
    P_{\text{AM}}(Y|X) + \alpha_{1} P_\text{LM}(Y) + \alpha_{2} |Y|
    \label{eq:lm}
\end{equation}
where $Y$ is the predicted sequence, $\alpha_1$ denote the weight of the language model and $\alpha_2$ is a word insertion penalty.
We Found that the examined models tend to be very confident regarding their predictions, i.e. the predicted probability distribution is massed for a single token at each timestep.

Figure~\ref{fig:logits} (upper) illustrate the top 4 predicted probabilities for a single word from a Librispeech test-clean sample. It is worth noting that the predicted probability distribution at the 5-th timestep is massed almost entirely at token "$E$". This outcome is sub-optimal, disrupting beam search effectiveness even when using a large LM weight $\alpha_1$.

\section{Proposed Approach}
As described earlier, the examined models generate predictions by applying $lm\_head$ on the emissions of the top transformer layer as described in Equation~\eqref{eq:baseline_logits}.

\begin{figure}[ht]
  \centering
  \includegraphics[trim=0 0 0 0, clip, width=\linewidth]{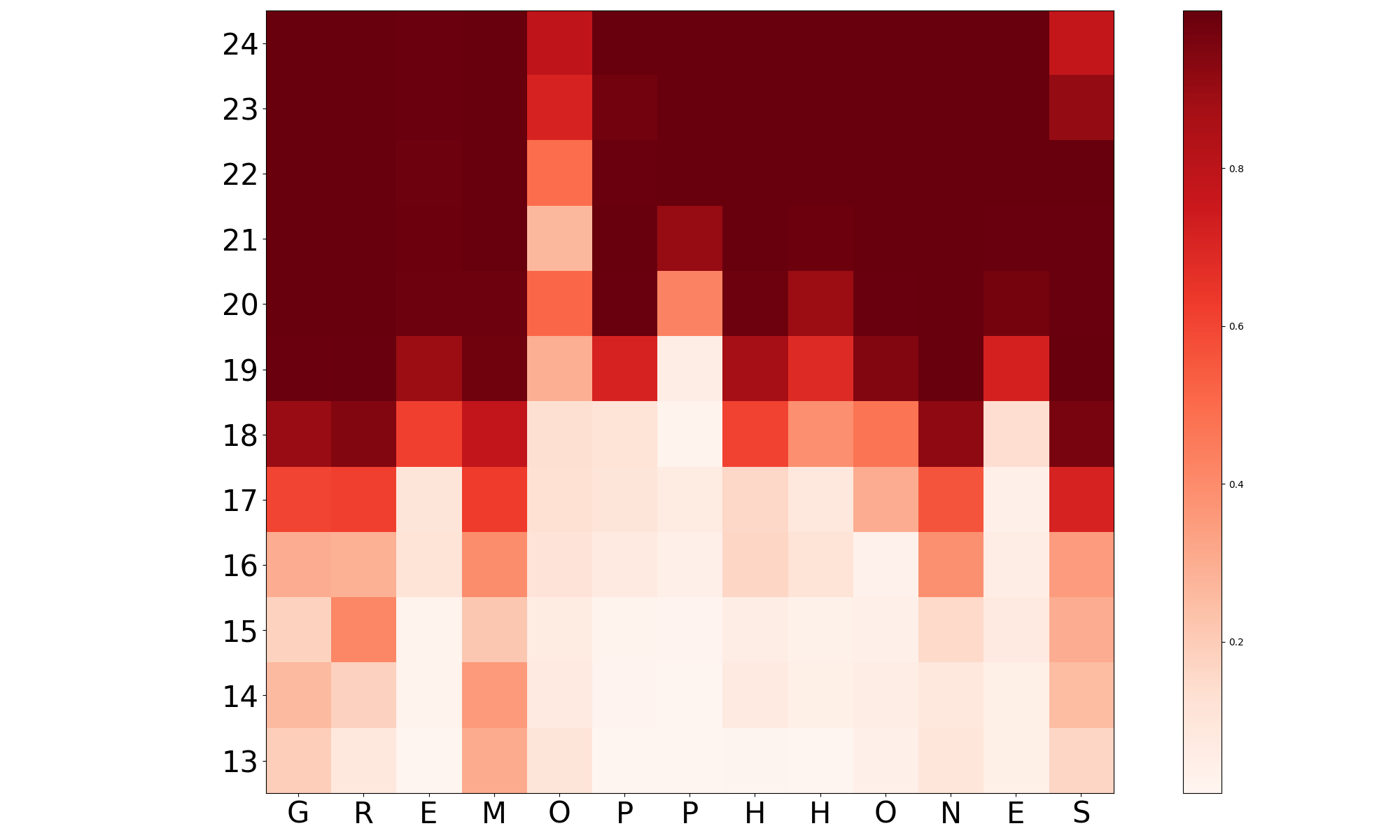}
    \caption{The confidence evolution throughout the top 12 transformer layers of Wav2vec 2.0 Large fine-tuned on 960 hours of Librispeech. Representations produced by the top (24th) layer are traditionally used for prediction. The prediction confidence increases at the top layers, ultimately reaching exceptionally high confidence levels.}
  \label{fig:evolution}
\end{figure}

\begin{figure}[ht]
  \centering
  \includegraphics[width=\linewidth]{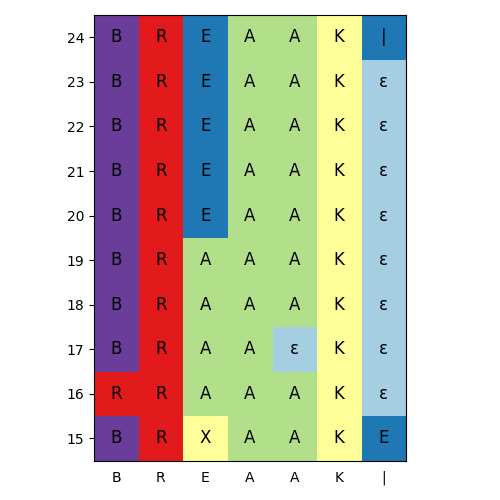}
    \caption{Predicted token evolution across top 10 transformer layers, generated using Wav2vec 2.0 Large fine-tuned on 960 hours of Librispeech. By using the logits of the top layer (24th), beam search with LM wrongly predicts "break" instead of "brake". The proposed approach managed to correctly predict "brake".}
  \label{fig:token_evolution}
\end{figure}

\begin{figure}[t]
  \centering
  \includegraphics[height=50mm,width=\linewidth]{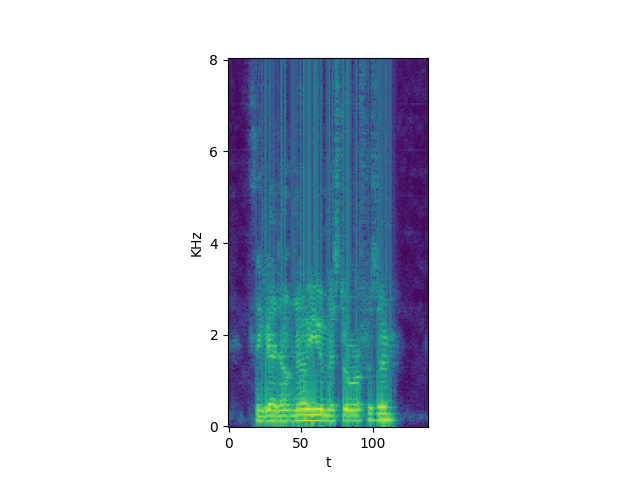}
    \caption{Log-spectogram of the audio processed in Figure~\ref{fig:attentions}. The first and last chunks of frames show no speech activity. }
  \label{fig:logspec}
\end{figure}

\begin{figure}[ht]
  \centering
  \includegraphics[trim=0 150 0 80, clip, width=\linewidth]{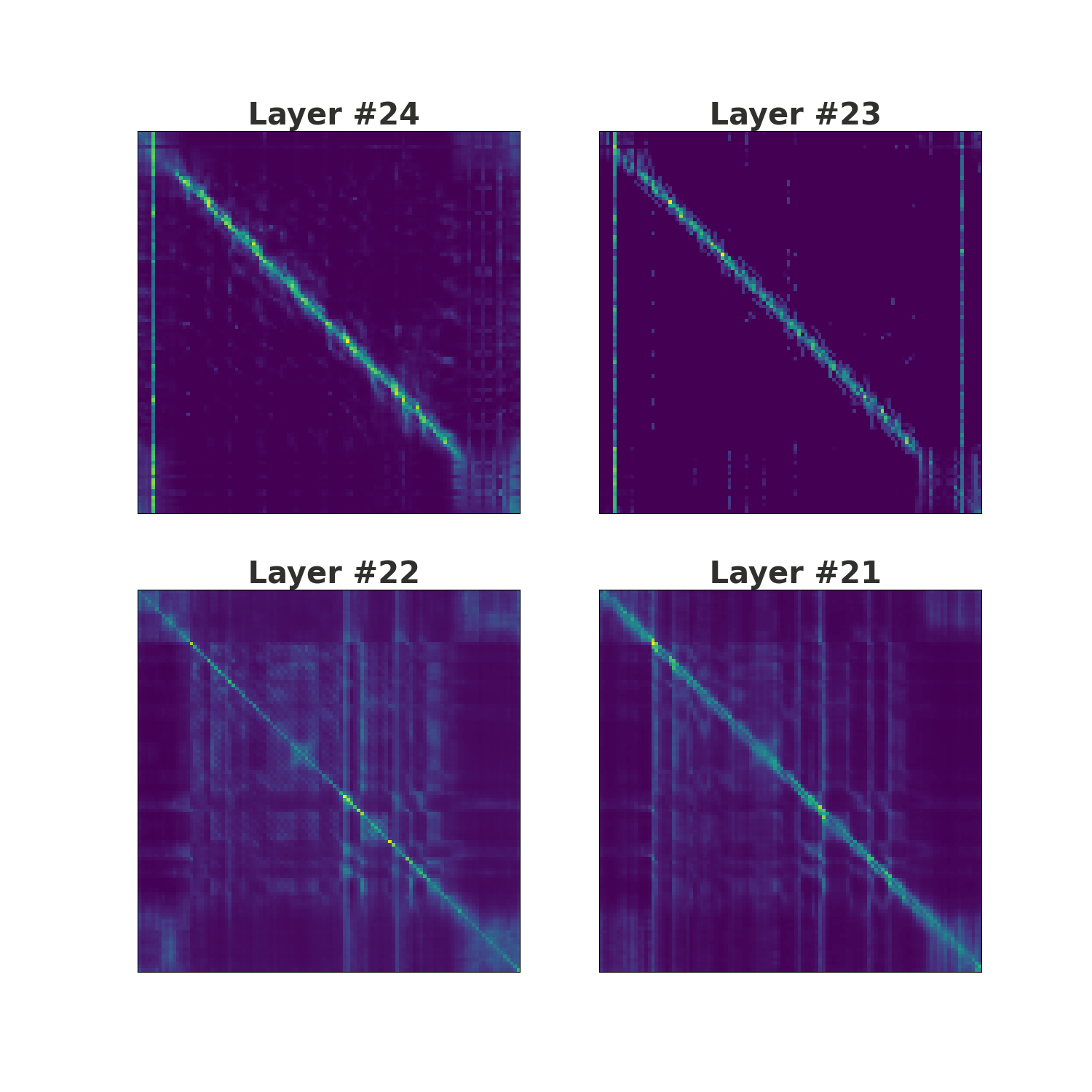}
    \caption{HuBERT-Large attention maps of the top four layers.\\ The attention weights are massed locally at the top layers, suggesting that the representations they produce rely on local information.}
  \label{fig:attentions}
\end{figure}

\begin{equation}
  logits(x_t) =lm\_head(H_N^t)
  \label{eq:baseline_logits}
\end{equation}

We empirically evaluated the models' confidence evolution throughout the transformer layers by employing $lm\_head$, followed by a Softmax operation, on the representations emitted by each layer. Our findings confirm that there is a confidence increase with each transformer layer, as shown in Figure~\ref{fig:evolution}.
We hypothesize that excessive confidence might prevent beam search from considering less confident predictions. Hence, we wish to smooth these highly confident logits as well as to directly harness attributes encoded within intermediate transformer layers~\cite{pasad2021layer}.

A use-case example is illustrated in Figure~\ref{fig:token_evolution}: for each timestep, we project intermediate representations using $lm\_head$ and choose the token with the largest logit at each layer. The illustration shows that a correct prediction may live in intermediate layers, rather than the top layer. Thus, utilizing intermediate layers can provide beam search with valuable alternatives. 

We commenced our experiments with traditional confidence reduction techniques, namely, label smoothing~\cite{szegedy2016rethinking} and temperature scaling. We fine-tuned Wav2vec 2.0 Large on 960 hours of Librispeech while employing label smoothing using different smoothing levels. The obtained performance was significantly inferior compared to the baseline model. Temperature scaling was indeed effective for reducing model confidence, however, we did not notice any performance increase. Also, temperature scaling does not take into account intermediate layers which may provide useful information.  

Based on this observation, we propose aggregating the logits of the top $M<N$ transformer layers, forming a relaxed and perhaps better informed logits to be used for prediction. As a consequence of the confidence increase across layers, a scaling mechanism is required to prevent the top layers from dominating the aggregation product. For this reason, We apply an $L2$ normalization on each layer's logits at each timestep, regularizing the impact of the more confident transformer layers on the aggregation product.

Formally, at each timestep our aggregation method can be described as follows:

\begin{equation}
  aggregated\_logits(X) = \sum_{n=N-M}^{N} lm\_head(\frac{H_n}{||H_n||_2})
  \label{eq:aggregation}
\end{equation}

\begin{table*}[ht]
  \begin{tabular}{ cccccccccccccc}
    \toprule
    \multicolumn{2}{c}{} & \multicolumn{2}{c}{} & \multicolumn{2}{c}{} & \multicolumn{4}{c}{Baseline} & \multicolumn{4}{c}{Layer Aggregation}\\
    \multicolumn{2}{c}{Model} & \multicolumn{2}{c}{Agg. coeff. ($\beta$)} & \multicolumn{2}{c}{\# Agg. Layers} & \multicolumn{2}{c}{test-clean} & \multicolumn{2}{c}{test-other} & \multicolumn{2}{c}{test-clean} & \multicolumn{2}{c}{test-other}\\\cmidrule(lr){7-8}\cmidrule(lr){9-10}\cmidrule(lr){11-12}\cmidrule(lr){13-14}
    {} & {} & {} & {} & {} & {} & {WER} & {CER} & {WER} & {CER} & {WER} & {CER} & {WER} & {CER}\\
    \midrule
    \multicolumn{2}{c}{Wav2vec 2.0 BASE} & \multicolumn{2}{c}{0.75} & \multicolumn{2}{c}{4} & 2.5 & 0.76 & 6.3 & 2.5 & \textbf{2.4} & \textbf{0.75} & \textbf{6.1} & \textbf{2.4}\\
    \multicolumn{2}{c}{Wav2vec 2.0 LARGE} & \multicolumn{2}{c}{0.85} & \multicolumn{2}{c}{12} & \textbf{2.1} & 0.64 & 5.0 & 2.0 & \textbf{2.1} & \textbf{0.63} & \textbf{4.9} & \textbf{1.9}\\
    \multicolumn{2}{c}{HUBERT LARGE} & \multicolumn{2}{c}{0.75} & \multicolumn{2}{c}{18} & \textbf{1.7} & \textbf{0.4} & 3.5 & 1.2 & \textbf{1.7} & \textbf{0.4} & \textbf{3.4} & \textbf{1.1}\\
    \multicolumn{2}{c}{HUBERT X-LARGE} & \multicolumn{2}{c}{0.9} & \multicolumn{2}{c}{24} & \textbf{1.6} & \textbf{0.4} & 3.0 & 1.0 & \textbf{1.6} & \textbf{0.4} & \textbf{2.9} & \textbf{0.9}\\
    \bottomrule
  \end{tabular}
  \begin{tabular}{ cccccccccccccc}
    \toprule
     \multicolumn{2}{c}{} & \multicolumn{2}{c}{} & \multicolumn{2}{c}{} & \multicolumn{4}{c}{Baseline} & \multicolumn{4}{c}{Layer Aggregation}\\
    \multicolumn{2}{c}{Model} & \multicolumn{2}{c}{Agg. coeff. ($\beta$)} & \multicolumn{2}{c}{\# Agg. Layers} & \multicolumn{2}{c}{dev-clean} & \multicolumn{2}{c}{dev-other} & \multicolumn{2}{c}{dev-clean} & \multicolumn{2}{c}{dev-other}\\\cmidrule(lr){7-8}\cmidrule(lr){9-10}\cmidrule(lr){11-12}\cmidrule(lr){13-14}
    {} & {} & {} & {} & {} & {} & {WER} & {CER} & {WER} & {CER} & {WER} & {CER} & {WER} & {CER}\\
    \midrule
    \multicolumn{2}{c}{Wav2vec 2.0 BASE} & \multicolumn{2}{c}{0.75} & \multicolumn{2}{c}{4} & 3.3 & 1.1 & 8.7 & 3.9 & \textbf{3.1} & \textbf{1.0} & \textbf{8.2} & \textbf{3.7}\\
    \multicolumn{2}{c}{Wav2vec 2.0 LARGE} & \multicolumn{2}{c}{0.85} & \multicolumn{2}{c}{12} & 2.0 & \textbf{0.6} & \textbf{4.7} & \textbf{1.9} & \textbf{1.9} & \textbf{0.6} & \textbf{4.7} & \textbf{1.9}\\
    \multicolumn{2}{c}{HUBERT LARGE} & \multicolumn{2}{c}{0.75} & \multicolumn{2}{c}{18} & \textbf{1.5} & \textbf{0.4} & 3.2 & 1.2 & \textbf{1.5} & \textbf{0.4} & \textbf{3.1} & \textbf{1.1}\\
    \multicolumn{2}{c}{HUBERT X-LARGE} & \multicolumn{2}{c}{0.9} & \multicolumn{2}{c}{24} & 1.7 & 0.4 & \textbf{2.6} & \textbf{1.0} & \textbf{1.5} & \textbf{0.3} & \textbf{2.6 }& \textbf{1.0}\\
    \bottomrule
  \end{tabular}
  \caption{Baseline and Layer aggregation results on Librispeech test and dev sets.}
  \label{tab:results}
\end{table*}

Finally, we interpolate our proposed aggregated logits with the logits produced by the top transformer layer:

\begin{equation}
  \beta \cdot logits(X) + (1-\beta) \cdot aggregated\_logits(X)
  \label{eq:interpolate}
\end{equation}

We hypothesize that using the layer-aggregated logits will assist with generating better transcriptions.

\section{Experimental Setup}
To properly explore to what degree the models are confident with their predictions, and to assess the impact of our proposed method, we experiment with different SSL speech recognition models of different sizes.

Specifically, we use Wav2vec 2.0 and HuBERT pre-trained speech recognition models. The size of the experimented models vary between ~95M and 964M parameters. All models were fine-tuned on 960 hours of Librispeech, while the HuBERT models were pre-trained on 60k hours of Libri-Light~\cite{kahn2020libri} and Wav2vec 2.0 models were fine-tuned on 960 hours of LibriSpeech~\cite{panayotov2015librispeech}. All experiments were conducted using models downloaded from the HuggingFace\footnote{All models are available at https://huggingface.co/models} platform. 

The reported experiments results were performed using Librispeech test and dev sets. We use the pyctcdecode\footnote{https://github.com/kensho-technologies/pyctcdecode} beam search decoder with a 4-gram LM. The decoder parameters were searched and optimized using the Ax toolkit\footnote{https://github.com/facebook/Ax}.
We tune the aggregation hyperparameters, namely, the number of aggregated layers ($N$) and aggregation tradeoff coefficient ($\beta$) on the dev-clean sets. 

We conduct two experimental setups: Baseline and Layer Aggregation. In the Baseline setup, we perform inference using the pre-trained and fine-tuned models and report the results obtained using beam search with a 4-gram LM. In the Layer-Aggregation setup, we employ our proposed approach and utilize the same 4-gram LM. Also, we perform a weight attention analysis by averaging the attention weights of all self-attention heads in a particular transformer layer.

Note, that  a 4-gram LM was utilized in all our experiments for simplicity and reproducibility sake, however, our method is not limited to a single LM and better results may possibly be obtained using a different high performing language model.

\section{Results}
Table~\ref{tab:results} reports the Baseline and proposed Layer Aggregation results on test and dev sets (both 'clean' and 'other' sets). The results indicate that in most cases layer aggregated logits either matches or improves WER and CER in both experimented models and in all model sizes.

By Aggregating the transformer layers logits we were able to relax highly confident predictions and as a consequence provide beam search a more diverse pool of alternatives (Figure~\ref{fig:logits}, bottom). For example, when transcribing: \emph{"There's a rocket waiting to transship you to the moon on the way to mercury right now gordon sighed"}, Wav2Vec 2.0 Large mistakenly replaced \emph{transship} with \emph{tranship}, while the proposed approach was able to fix that error.

Although the Baseline method included the correct transcription as one of its candidates, the probability it produced was less than 15\% for an extra \emph{s} resulting with a wrong transcription despite the fact that this candidate received higher LM score. Using aggregated logits, we were was able to increase that probability, assisting beam search with tracking the correct transcription.

In addition, we noticed that decoding using the proposed approach opens the door for transcribing relatively rare words which were mistakenly predicted by the Baseline approach. For example \emph{"Phedrus"} was corrected to \emph{"Phaedrus"}, \emph{"Credius"} corrected to \emph{"Critias"}, \emph{"Chelsey"} corrected to \emph{"Chelsea"}. 

Furthermore, we examined the attention weights computed by the top layers of the transformer. For that, an audio sample depicted in Figure~\ref{fig:logspec} was processed. Interestingly, we observed that the attention weights are locally concentrated at the top layers (Figure~\ref{fig:attentions}) while the attention magnitude decreases when descending down to intermediate layers. This observation provide a different angle explaining our hypothesize, and further validate the findings from previous works that claimed the top layers tend to focus on local features. Noteworthy, the no-speech-activity frames provides additional explanation to the attention weights behavior at the beginning and ending of the processed sequence.

\section{Consclusions}
In this work, we explored and analyzed the prediction flow of state-of-the-art speech recognition systems. We found that these models generate extremely confident predictions. We assessed the impact of this property on a beam search decoder with an integrated LM, which is widely used in speech recognition frameworks.
We analyzed and visualized the token prediction evolution process throughout the transformer layers, and used our findings to design a layer aggregation method which is specifically suited for beam search decoding. Our suggested method, which consist of aggregating logits from deeper intermediate layers, was able to reduce WER and CER when applied to Wav2vec 2.0 and HuBERT models.

\bibliographystyle{IEEEtran}

\bibliography{template}


\end{document}